\documentclass{article} 
\usepackage{iclr2021_conference,times}

\usepackage{amsmath,amsfonts,bm}

\def\eqref#1{equation~\ref{#1}}

\def\1{\bm{1}}

\DeclareMathAlphabet{\mathsfit}{\encodingdefault}{\sfdefault}{m}{sl}
\SetMathAlphabet{\mathsfit}{bold}{\encodingdefault}{\sfdefault}{bx}{n}

\usepackage{blindtext}
\usepackage[hyphens]{url}
\usepackage{hyperref}
\usepackage{url}

\title{Perspectives on machine learning from \\ psychology's reproducibility crisis}

\author{Samuel J. Bell \\
Department of Computer Science and Technology\\
University of Cambridge \\
\texttt{sjb326@cam.ac.uk} \\
\And
Onno P. Kampman \\
Department of Psychology \\
University of Cambridge \\
\texttt{opk20@cam.ac.uk} \\
}

\iclrfinalcopy 
\begin{document}

\maketitle

\begin{abstract}

In the early 2010s, a crisis of reproducibility rocked the field of psychology.
Following a period of reflection, the field has responded with radical reform of its scientific practices.
More recently, similar questions about the reproducibility of machine learning research have also come to the fore.
In this short paper, we present select ideas from psychology's reformation, translating them into relevance for a machine learning audience.

\end{abstract}

\section{Introduction}

The machine learning community spans a broad spectrum of priorities.
Those of us that lean toward engineering may focus on developing effective models and deploying systems that are practical and scalable.
Those that lean toward science may be more concerned with investigating and understanding empirical phenomena, or devising new theoretical accounts.
Though at first glance these aims seem diverse, they share a common core: findings that are valid, robust and generalizable.
When results don't stand up to such scrutiny, researchers build their investigations on foundations of sand, and engineers waste their efforts striving for unattainable benchmarks.
Worse, systems that do reach deployment may function in unanticipated ways, with significant consequences in areas such as medicine or law.

In psychology (broadly defined here from social and experimental psychology to cognitive neuroscience), the \emph{reproducibility crisis} has inflicted long-lasting reputational damage~\citep{Pashler2012}.
Contributing factors include selecting analyses to find positive results~\citep{Simmons2011, Wagenmakers2012}; partial reporting~\citep{Masicampo2012}; publication bias and perverse incentive structures~\citep{Nosek2012}; and questionable research practices~\citep{John2012}.
In response, the field has undergone significant reform and implemented a variety of new practices, some of which are useful case studies for machine learning.

In machine learning, reproducibility is increasingly a focus of attention~\citep{Sculley2018, Gundersen2018, Forde2019, Lipton2019, Bouthillier2019}.
In many ways, our community has excellent practices, such as public code and data, open-access publications, and reproducibility checklists~\citep{Pineau2020}.
Even the propensity for shared tasks and benchmarks, though not without criticism~\citep{Sculley2018}, at least provides a level playing field for comparison.
However, while we note positive practices for \textit{methodological} reproducibility, generalizable and useful results require \textit{inferential} reproducibility~\citep{Gundersen2018}.
On this spectrum, methodological reproducibility verifies a \emph{result} using as similar methods as possible (e.g.\ using the original code), where inferential reproducibility verifies a \emph{conclusion}, using independent methods to test its core ideas~\citep{Goodman2018}.

Psychology provides a particularly useful comparison for machine learning: both fields study complex systems with surprising emergent behaviors, whether those systems are human minds or deep neural networks.
However, it is worth noting the limitations of the analogy too.
Psychology is an empirical and theoretical discipline, albeit with potential downstream applications in psychiatry and medicine.
By contrast, machine learning---even at its most scientific---is fundamentally an applied discipline.
In this short position paper, we survey psychology's reinvention as a more robust discipline.
We set out novel suggestions for machine learning research practices, with a particular focus on the sub-field of deep learning.

\section{Methodological changes}
\label{sec:robust-statistics}

Statistical hypothesis testing is an essential component of the scientific method, and should be used more widely in machine learning.
However, psychologists have identified certain issues prevalent in their use of statistics, many of which are also relevant to our discipline.

\subsection{A priori hypotheses}

A key issue is HARKing (Hypothesizing After Results are Known), where a hypothesis is devised \emph{after} the results have been analyzed~\citep{Kerr1998}.
In machine learning, HARKing may take the form of testing various ideas against a benchmark, and only selecting the most promising for publication.
It may also apply to our choice of performance metrics, or to our choice of problem dataset.

HARKing comes in various flavors, though all invalidate any use of statistical hypothesis testing.
\citet{Kerr1998} puts the problem succinctly, stating that HARKing ``translates Type 1 errors into theory''.
To counter this, researchers should declare clearly-motivated hypotheses, alongside falsifiable predictions, before experimentation.
While this recommendation may not seem relevant to those deploying machine learning in practice, this is precisely when the scientific method is at its most valuable.
Deployed models always have real world consequences, but clear hypotheses can prevent moving goalposts, letting us better understand the effects of our systems.

\subsection{Predefined methods and preregistration}

Type 1 errors also arise from the wide range of choices available for data analysis~\citep{Simmons2011}.
Plunging repeatedly into the data to search for a result, commonly known as ``\(p\)-hacking'', invalidates the core assumption of a statistical hypothesis test: the probability that a \emph{single} result is due to chance~\citep{Kriegeskorte2009}.
With a suitably large search space, false positive results are almost certain to emerge.

In deep learning, this issue occurs up when searching over model hyperparameters to find a desired result.
As a starting point, hypotheses could clearly state the search space of hyperparameters under consideration.
These hyperparameters could even be considered \emph{auxiliary} hypotheses about acceptable bounds.
One approach to support this is \emph{preregistration}~\citep{Nosek2012, Wagenmakers2012}.
With preregistration, researchers publicly declare their hypotheses, experiments, and analysis plans before running their experiments.
The final paper can then reference the preregistration, as a foil against practices such as HARKing, \(p\)-hacking, and selectively reporting only experiments with desirable results.
Machine learning is already experimenting with preregistration, including several preregistered-only workshops at major conferences.

Even in the psychological sciences, whether preregistration can apply to computational research and methods development is debated~\citep{MacEachern2019}.
Taking our example of hyperparameter search, it may be challenging to \emph{a priori} predict what is often considered a problem of exploration.
Alternatively, the preregistration might state the selection methodology or tuning algorithm, or the researcher may conduct a series of preliminary experiments to develop their intuition.
In this case, it is crucial that such exploratory work is clearly delineated from any confirmatory hypothesis testing~\citep{Wagenmakers2012}.

\subsection{Multiverse analyses}

Even with predefined hypotheses and methods, results are still sensitive to a researcher's choices.
A multiverse analysis~\citep{Steegen2016} aims to address this.
For every decision taken regarding how to analyze results, a different choice could have taken its place~\citep{gelman2013}.
In effect, this produces a decision tree of result sets, with a unique set at each leaf.
Depending on our chosen leaf, we may reach different conclusions.
A result that holds for only select leaves cannot be considered robust.

Even if our decisions are preregistered, our conclusions remain limited to a single leaf, and insufficient to understand sensitivity to different choices.
By computing a multiverse of results, we can begin to understand under which circumstances our conclusions may hold.

Unsurprisingly, we see this idea as applicable to hyperparameter analysis.
Often, reported ``gains'' are the result of hyperparameter changes, rather than the claims of the authors~\citep{Lipton2019, Henderson2017, Melis2018, Reimers2017, Bouthillier2019}.
A common suggestion is to report both the selection methodology and the end results~\citep{Sculley2018}.
However, though this may increase confidence in the result's validity, it does not get us generality.
We want to understand whether our conclusions hold \textit{in general}, or only \textit{in specific instances}.
Our proposal would consider the multiverse of model specifications, explicitly reporting results for all reasonable hyperparameter choices.
This approach is similar in spirit to \citet{Reimers2017}, who report the distribution of results over random seeds, or to \citet{Dodge2020}, who sample hyperparameter space to estimate expected maximum performance.

\section{Cultural changes}
\label{sec:culture}

Beyond methodology, the crisis in psychology has ushered in a period of reflection about cultural norms.
In machine learning, similar cultural changes could also emerge.
Though not every proposal in this section was entirely new to psychologists, many have seen more widespread acceptance.

\subsection{Publishing and registered reports}
\label{sec:registered-reports}

Though machine learning has adopted open-access publishing, other publication practices uphold questionable incentive structures.
Publication bias---where positive results are more likely to be published---is a key issue for both psychology~\citep{Nosek2012, Masicampo2012} and machine learning~\citep{Pineau2020}.
Not only can negative results be informative, but running experiments that have been tried before yet never published is far from efficient.
The quest for a positive result is also a strong incentive for questionable research practices. 
A meta-analysis in psychology found an unexpected number of papers reporting results suspiciously \textit{just} below the \(p=0.05\) significance threshold~\citep{Masicampo2012}.

The adoption of \emph{registered reports}~\citep{chambers2013registered} in psychology has been a useful tool to counter publication bias and increase publication of null findings~\citep{Allen2019}.
Here, a paper's motivations, hypotheses, and methods are submitted for review, before any experiments are performed.
Publication decisions are based solely on the value of the research question and the rigor of the methods, and accepted papers will be published regardless of a positive or negative result.
Registered reports can also provide early feedback to researchers.
In machine learning, such feedback should be of particular interest in light of mounting compute costs.

\subsection{Incentives}

The incentive structure of research careers, despite our best intentions, can sometimes lead to questionable research practices.
\citet{John2012} surveyed over 2,000 psychologists to measure the prevalence of such practices, with striking results: 67\% admit to selective reporting, 54\% to HARKing.
\emph{True} prevalence estimates are much higher across a host of practices.
Such a study could easily be conducted in machine learning, building an understanding of our practices.

With this information, we can implement subtle shifts in incentives.
For example, certain psychology venues (e.g.\ the Journal of Experimental Psychology) reject papers that fail to meet reproducibility standards.
We praise the machine learning reproducibility checklists introduced recently~\citep{Gundersen2018, Pineau2020}, though note that such requirements are never binding.

A surprisingly effective practice is awarding \emph{badges} to reproducible research.
This badge is published with the paper to highlight good practice, such as public code or data.
One journal reports that badges led to a rise from 3\% to 39\% in papers making their data publicly-available~\citep{kidwell2016badges}.
In a conference setting, a reproducibility award could sit alongside the coveted ``Best Paper''.
Such incentives could provide an additional route to credibility within the machine learning community and beyond.

\subsection{Communication}

A compelling narrative may seem like a good idea to improve one's reputation and the reach of one's work.
Yet, forcing empirical results into a neat storyline can distort the evidence, encouraging claims that do not stand up to scrutiny~\citep{Katz2013}.
Psychology is full of such cautionary tales, including claims that striking a powerful pose builds confidence~\citep{carney2010power}; that extra-sensory perception is possible~\citep{bem2011feeling}; and that willpower is a finite resource to be depleted~\citep{baumeister1998ego}.
These findings have each been debunked, though many are still believed to be true among the general public.
In spite of a focus on ``wins''~\citep{Sculley2018} and being ``state-of-the-art'', the machine learning community would do well to heed the warnings from such examples.

\subsection{Collaboration}

While competitive practices are the norm in machine learning, psychology has benefited from a collaborative response to its crisis.
\emph{Many Labs}~\citep{Klein2014} is a series of large-scale replication studies of influential psychological effects, conducted by multiple different labs.
In machine learning, a similar movement should make it easy to participate in replications, and allow for a critical assessment of reproducibility.
Conferences might consider a dedicated replications track.

Crucially, these replications were conducted with neither the involvement of the original authors, nor with access to their materials.
By attempting to replicate solely from the published paper, we can hone in on the \emph{inferential} reproducibility of the claims.
A similar attempt in machine learning would aim to reproduce key findings from our community, though without access to the paper's code.

\section{Beyond reproducibility}
\label{sec:theory}

Important as empirical rigor is, we may also reflect on the very nature and limits of knowledge and theory in our discipline.
In recent years, theoreticians have taken a more central role in psychology's crisis~\citep{Oberauer2019}.
They argue that reproducibility issues in psychology are not just the result of shaky methods, but also derive from the inherent difficulty of developing good theory~\citep{Szollosi2021}.
Many theories in psychology can explain phenomena, but fail to make concrete falsifiable predictions~\citep{Borsboom2013}.
Thus, conflicting theories often co-exist, making it hard to develop cumulative theory over time~\citep{eronen2021theory}.

Formulating strong theory is also a challenge in machine learning.
Theory is desirable because it allows us to understand, predict and reason about unobserved model behavior.
For example, while there is a growing body of work that characterizes the properties of simple neural networks, these findings rarely generalize to the more complex models used in practice.
These larger models are typically understood primarily through empirical investigation.
However, several authors~\citep{Lipton2019, Bouthillier2019} warn against conflating intuition and experience with theory, such that intuitions come to be regarded as fact.
Inspired by physics, \citet{Forde2019} suggest a bottom-up approach to developing theoretical accounts of complex models.
The analogy in psychology might be developing theories of human cognition from the study of neurons and circuitry.
In contrast, we suggest that frameworks from the psychological sciences can help us develop useful \emph{top-down} theories of behavior and mechanism~\citep{cummins2000does}, even when lacking a first principles understanding.

In this survey, several ideas from psychology are presented as inspiration for methodological reform, cultural change, and new opportunities for theoretical growth.
We hope they prove useful in promoting a more robust machine learning discipline. 

\subsubsection*{Acknowledgments}
This paper was inspired by Amy Orben's excellent lecture series, \emph{Psychology as a Robust Science}. We would also like to thank Neil Lawrence and Creighton Heaukulani for their helpful feedback.

\newpage
\bibliography{iclr2021_conference}

\begin{thebibliography}{35}
\providecommand{\natexlab}[1]{#1}
\providecommand{\url}[1]{\texttt{#1}}
\expandafter\ifx\csname urlstyle\endcsname\relax
  \providecommand{\doi}[1]{doi: #1}\else
  \providecommand{\doi}{doi: \begingroup \urlstyle{rm}\Url}\fi

\bibitem[Allen \& Mehler(2019)Allen and Mehler]{Allen2019}
Christopher Allen and David~M.A. Mehler.
\newblock Open science challenges, benefits and tips in early career and
  beyond.
\newblock \emph{PLoS Biology}, 17\penalty0 (5), 2019.
\newblock \doi{10.1371/journal.pbio.3000246}.

\bibitem[Baumeister et~al.(1998)Baumeister, Bratslavsky, Muraven, and
  Tice]{baumeister1998ego}
Roy~F. Baumeister, Ellen Bratslavsky, Mark Muraven, and Dianne~M. Tice.
\newblock Ego depletion: Is the active self a limited resource?
\newblock \emph{Journal of Personality and Social Psychology}, 74\penalty0
  (5):\penalty0 1252--1265, 1998.

\bibitem[Bem(2011)]{bem2011feeling}
Daryl~J. Bem.
\newblock Feeling the future: Experimental evidence for anomalous retroactive
  influences on cognition and affect.
\newblock \emph{Journal of Personality and Social Psychology}, 100\penalty0
  (3):\penalty0 407--425, 2011.

\bibitem[Borsboom(2013)]{Borsboom2013}
Denny Borsboom.
\newblock Theoretical amnesia., 2013.
\newblock
  \url{http://osc.centerforopenscience.org/2013/11/20/theoretical-amnesia}.

\bibitem[Bouthillier et~al.(2019)Bouthillier, Laurent, and
  Vincent]{Bouthillier2019}
Xavier Bouthillier, C{\'{e}}sar Laurent, and Pascal Vincent.
\newblock Unreproducible research is reproducible.
\newblock \emph{36th International Conference on Machine Learning, ICML 2019},
  pp.\  1150--1159, 2019.

\bibitem[Carney et~al.(2010)Carney, Cuddy, and Yap]{carney2010power}
Dana~R. Carney, Amy~J.C. Cuddy, and Andy~J. Yap.
\newblock Power posing: Brief nonverbal displays affect neuroendocrine levels
  and risk tolerance.
\newblock \emph{Psychological Science}, 21\penalty0 (10):\penalty0 1363--1368,
  2010.

\bibitem[Chambers(2013)]{chambers2013registered}
Christopher~D. Chambers.
\newblock Registered reports: A new publishing initiative at {Cortex}.
\newblock \emph{Cortex}, 49\penalty0 (3):\penalty0 609--610, 2013.

\bibitem[Cummins(2000)]{cummins2000does}
Robert Cummins.
\newblock ``{How} does it work?'' versus ``{What} are the laws?'': Two
  conceptions of psychological explanation.
\newblock In \emph{Explanation and cognition}, pp.\  117--144. {MIT Press},
  2000.

\bibitem[Dodge et~al.(2019)Dodge, Gururangan, Card, Schwartz, and
  Smith]{Dodge2020}
Jesse Dodge, Suchin Gururangan, Dallas Card, Roy Schwartz, and Noah~A. Smith.
\newblock Show your work: Improved reporting of experimental results.
\newblock In \emph{Proceedings of the 2019 Conference on Empirical Methods in
  Natural Language Processing and the 9th International Joint Conference on
  Natural Language Processing, EMNLP-IJCNLP}, pp.\  2185--2194, 2019.
\newblock \doi{10.18653/v1/D19-1224}.

\bibitem[Eronen \& Bringmann(2021)Eronen and Bringmann]{eronen2021theory}
Markus~I. Eronen and Laura~F. Bringmann.
\newblock The theory crisis in psychology: How to move forward.
\newblock \emph{Perspectives on Psychological Science}, 2021.
\newblock \doi{10.1177/1745691620970586}.

\bibitem[Forde \& Paganini(2019)Forde and Paganini]{Forde2019}
Jessica~Zosa Forde and Michela Paganini.
\newblock The scientific method in the science of machine learning, 2019.
\newblock URL \url{https://arxiv.org/abs/1904.10922}.

\bibitem[Gelman \& Loken(2013)Gelman and Loken]{gelman2013}
Andrew Gelman and Eric Loken.
\newblock The garden of forking paths: Why multiple comparisons can be a
  problem, even when there is no “fishing expedition” or “p-hacking”
  and the research hypothesis was posited ahead of time.
\newblock \emph{Unpublished manuscript}, 2013.
\newblock URL
  \url{http://www.stat.columbia.edu/~gelman/research/unpublished/p_hacking.pdf}.

\bibitem[Goodman et~al.(2018)Goodman, Fanelli, and Ioannidis]{Goodman2018}
Steven~N. Goodman, Daniele Fanelli, and John~P.A. Ioannidis.
\newblock What does research reproducibility mean?
\newblock \emph{Science Translational Medicine}, 8\penalty0 (341):\penalty0
  96--102, 2018.
\newblock \doi{10.1126/scitranslmed.aaf5027}.

\bibitem[Gundersen \& Kjensmo(2018)Gundersen and Kjensmo]{Gundersen2018}
Odd~Erik Gundersen and Sigbj{\o}rn Kjensmo.
\newblock State of the art: Reproducibility in artificial intelligence.
\newblock In \emph{Proceedings of the AAAI Conference on Artificial
  Intelligence}, volume~32, 2018.

\bibitem[Henderson et~al.(2018)Henderson, Islam, Bachman, Pineau, Precup, and
  Meger]{Henderson2017}
Peter Henderson, Riashat Islam, Philip Bachman, Joelle Pineau, Doina Precup,
  and David Meger.
\newblock Deep reinforcement learning that matters.
\newblock In \emph{Proceedings of the AAAI Conference on Artificial
  Intelligence}, volume~32, 2018.

\bibitem[John et~al.(2012)John, Loewenstein, and Prelec]{John2012}
Leslie~K. John, George Loewenstein, and Drazen Prelec.
\newblock Measuring the prevalence of questionable research practices with
  incentives for truth telling.
\newblock \emph{Psychological Science}, 23\penalty0 (5):\penalty0 524--532,
  2012.
\newblock \doi{10.1177/0956797611430953}.

\bibitem[Katz(2013)]{Katz2013}
Yarden Katz.
\newblock Against storytelling of scientific results.
\newblock \emph{Nature Methods}, 10\penalty0 (11):\penalty0 1045, 2013.
\newblock \doi{10.1038/nmeth.2699}.

\bibitem[Kerr(1998)]{Kerr1998}
Norbert~L. Kerr.
\newblock {HARKing}: Hypothesizing after the results are known.
\newblock \emph{Personality and Social Psychology Review}, 2\penalty0
  (3):\penalty0 196--217, 1998.
\newblock \doi{10.1207/s15327957pspr0203_4}.

\bibitem[Kidwell et~al.(2016)Kidwell, Lazarevi{\'c}, Baranski, Hardwicke,
  Piechowski, Falkenberg, Kennett, Slowik, Sonnleitner, Hess-Holden,
  et~al.]{kidwell2016badges}
Mallory~C. Kidwell, Ljiljana~B. Lazarevi{\'c}, Erica Baranski, Tom~E.
  Hardwicke, Sarah Piechowski, Lina-Sophia Falkenberg, Curtis Kennett,
  Agnieszka Slowik, Carina Sonnleitner, Chelsey Hess-Holden, et~al.
\newblock Badges to acknowledge open practices: A simple, low-cost, effective
  method for increasing transparency.
\newblock \emph{PLoS biology}, 14\penalty0 (5):\penalty0 e1002456, 2016.

\bibitem[Klein et~al.(2014)Klein, Ratliff, Vianello, Adams, Bahn{\'{i}}k,
  Bernstein, Bocian, Brandt, Brooks, Brumbaugh, Cemalcilar, Chandler, Cheong,
  Davis, Devos, Eisner, Frankowska, Furrow, Galliani, Hasselman, Hicks,
  Hovermale, Hunt, Huntsinger, Ijzerman, John, Joy-Gaba, Kappes, Krueger,
  Kurtz, Levitan, Mallett, Morris, Nelson, Nier, Packard, Pilati, Rutchick,
  Schmidt, Skorinko, Smith, Steiner, Storbeck, {Van Swol}, Thompson, {Van 'T
  Veer}, Vaughn, Vranka, Wichman, Woodzicka, and Nosek]{Klein2014}
Richard~A. Klein, Kate~A. Ratliff, Michelangelo Vianello, Reginald~B. Adams,
  {\v{S}}t{\v{e}}p{\'{a}}n Bahn{\'{i}}k, Michael~J. Bernstein, Konrad Bocian,
  Mark~J. Brandt, Beach Brooks, Claudia~Chloe Brumbaugh, Zeynep Cemalcilar,
  Jesse Chandler, Winnee Cheong, William~E. Davis, Thierry Devos, Matthew
  Eisner, Natalia Frankowska, David Furrow, Elisa~Maria Galliani, Fred
  Hasselman, Joshua~A. Hicks, James~F. Hovermale, S.~Jane Hunt, Jeffrey~R.
  Huntsinger, Hans Ijzerman, Melissa~Sue John, Jennifer~A. Joy-Gaba,
  Heather~Barry Kappes, Lacy~E. Krueger, Jaime Kurtz, Carmel~A. Levitan,
  Robyn~K. Mallett, Wendy~L. Morris, Anthony~J. Nelson, Jason~A. Nier, Grant
  Packard, Ronaldo Pilati, Abraham~M. Rutchick, Kathleen Schmidt, Jeanine~L.
  Skorinko, Robert Smith, Troy~G. Steiner, Justin Storbeck, Lyn~M. {Van Swol},
  Donna Thompson, A.~E. {Van 'T Veer}, Leigh~Ann Vaughn, Marek Vranka, Aaron~L.
  Wichman, Julie~A. Woodzicka, and Brian~A. Nosek.
\newblock Investigating variation in replicability: A ``{Many Labs}''
  replication project.
\newblock \emph{Social Psychology}, 45\penalty0 (3):\penalty0 142--152, 2014.
\newblock \doi{10.1027/1864-9335/a000178}.

\bibitem[Kriegeskorte et~al.(2009)Kriegeskorte, Simmons, Bellgowan, and
  Baker]{Kriegeskorte2009}
Nikolaus Kriegeskorte, W.~Kyle Simmons, Patrick~S. Bellgowan, and Chris~I.
  Baker.
\newblock Circular analysis in systems neuroscience: The dangers of double
  dipping.
\newblock \emph{Nature Neuroscience}, 12\penalty0 (5):\penalty0 535--540, 2009.
\newblock \doi{10.1038/nn.2303}.

\bibitem[Lipton \& Steinhardt(2019)Lipton and Steinhardt]{Lipton2019}
Zachary~C. Lipton and Jacob Steinhardt.
\newblock Troubling trends in machine learning scholarship.
\newblock \emph{Queue}, 17\penalty0 (1):\penalty0 45–77, 2019.
\newblock \doi{10.1145/3317287.3328534}.

\bibitem[MacEachern \& {Van Zandt}(2019)MacEachern and {Van
  Zandt}]{MacEachern2019}
Steven~N. MacEachern and Trisha {Van Zandt}.
\newblock Preregistration of modeling exercises may not be useful.
\newblock \emph{Computational Brain \& Behavior}, 2\penalty0 (3-4):\penalty0
  179--182, 2019.
\newblock \doi{10.1007/s42113-019-00038-x}.

\bibitem[Masicampo \& Lalande(2012)Masicampo and Lalande]{Masicampo2012}
E.~J. Masicampo and Daniel~R. Lalande.
\newblock A peculiar prevalence of p values just below .05.
\newblock \emph{Quarterly Journal of Experimental Psychology}, 65\penalty0
  (11):\penalty0 2271--2279, 2012.
\newblock \doi{10.1080/17470218.2012.711335}.

\bibitem[Melis et~al.(2018)Melis, Dyer, and Blunsom]{Melis2018}
G{\'{a}}bor Melis, Chris Dyer, and Phil Blunsom.
\newblock On the state of the art of evaluation in neural language models.
\newblock \emph{6th International Conference on Learning Representations, ICLR
  2018 - Conference Track Proceedings}, pp.\  1--10, 2018.

\bibitem[Nosek et~al.(2012)Nosek, Spies, and Motyl]{Nosek2012}
Brian~A. Nosek, Jeffrey~R. Spies, and Matt Motyl.
\newblock Scientific utopia: {II.} {R}estructuring incentives and practices to
  promote truth over publishability.
\newblock \emph{Perspectives on Psychological Science}, 7\penalty0
  (6):\penalty0 615--631, 2012.
\newblock \doi{10.1177/1745691612459058}.

\bibitem[Oberauer \& Lewandowsky(2019)Oberauer and Lewandowsky]{Oberauer2019}
Klaus Oberauer and Stephan Lewandowsky.
\newblock Addressing the theory crisis in psychology.
\newblock \emph{Psychonomic Bulletin and Review}, 26\penalty0 (5):\penalty0
  1596--1618, 2019.
\newblock \doi{10.3758/s13423-019-01645-2}.

\bibitem[Pashler \& Wagenmakers(2012)Pashler and Wagenmakers]{Pashler2012}
Harold Pashler and Eric-Jan Wagenmakers.
\newblock Editors' introduction to the special section on replicability in
  psychological science: A crisis of confidence?
\newblock \emph{Perspectives on Psychological Science}, 7\penalty0
  (6):\penalty0 528--530, 2012.
\newblock \doi{10.1177/1745691612465253}.

\bibitem[Pineau et~al.(2020)Pineau, Vincent-Lamarre, Sinha, Larivi{\`{e}}re,
  Beygelzimer, D'Alch{\'{e}}-Buc, Fox, and Larochelle]{Pineau2020}
Joelle Pineau, Philippe Vincent-Lamarre, Koustuv Sinha, Vincent
  Larivi{\`{e}}re, Alina Beygelzimer, Florence D'Alch{\'{e}}-Buc, Emily Fox,
  and Hugo Larochelle.
\newblock Improving reproducibility in machine learning research (a report from
  the {NeurIPS} 2019 reproducibility program).
\newblock \emph{arXiv}, 2020.

\bibitem[Reimers \& Gurevych(2017)Reimers and Gurevych]{Reimers2017}
Nils Reimers and Iryna Gurevych.
\newblock Reporting score distributions makes a difference: Performance study
  of {LSTM}-networks for sequence tagging.
\newblock \emph{Proceedings of the 2017 Conference on Empirical Methods in
  Natural Language Processing, EMNLP 2017}, pp.\  338--348, 2017.
\newblock \doi{10.18653/v1/d17-1035}.

\bibitem[Sculley et~al.(2018)Sculley, Snoek, Rahimi, and
  Wiltschko]{Sculley2018}
D.~Sculley, Jasper Snoek, Ali Rahimi, and Alex Wiltschko.
\newblock Winner's curse? {On} pace, progress, and empirical rigor.
\newblock \emph{6th International Conference on Learning Representations, ICLR
  2018 - Workshop Track Proceedings}, 2018.

\bibitem[Simmons et~al.(2011)Simmons, Nelson, and Simonsohn]{Simmons2011}
Joseph~P. Simmons, Leif~D. Nelson, and Uri Simonsohn.
\newblock False-positive psychology: Undisclosed flexibility in data collection
  and analysis allows presenting anything as significant.
\newblock \emph{Psychological Science}, 22\penalty0 (11):\penalty0 1359--1366,
  2011.
\newblock \doi{10.1177/0956797611417632}.

\bibitem[Steegen et~al.(2016)Steegen, Tuerlinckx, Gelman, and
  Vanpaemel]{Steegen2016}
Sara Steegen, Francis Tuerlinckx, Andrew Gelman, and Wolf Vanpaemel.
\newblock Increasing transparency through a multiverse analysis.
\newblock \emph{Perspectives on Psychological Science}, 11\penalty0
  (5):\penalty0 702--712, 2016.
\newblock \doi{10.1177/1745691616658637}.

\bibitem[Szollosi \& Donkin(2021)Szollosi and Donkin]{Szollosi2021}
Aba Szollosi and Chris Donkin.
\newblock Arrested theory development: The misguided distinction between
  exploratory and confirmatory research.
\newblock \emph{Perspectives on Psychological Science}, 2021.
\newblock \doi{10.1177/1745691620966796}.

\bibitem[Wagenmakers et~al.(2012)Wagenmakers, Wetzels, Borsboom, van~der Maas,
  and Kievit]{Wagenmakers2012}
Eric-Jan Wagenmakers, Ruud Wetzels, Denny Borsboom, Han~L.J. van~der Maas, and
  Rogier~A. Kievit.
\newblock An agenda for purely confirmatory research.
\newblock \emph{Perspectives on Psychological Science}, 7\penalty0
  (6):\penalty0 632--638, 2012.
\newblock \doi{10.1177/1745691612463078}.

\end{thebibliography}
\bibliographystyle{iclr2021_conference}

\end{document}